\documentclass{article}

 \usepackage[main, final]{neurips_2026}

\usepackage[utf8]{inputenc} 
\usepackage[T1]{fontenc}    
\usepackage{hyperref}       
\usepackage{url}            
\usepackage{booktabs}       
\usepackage{amsfonts}       
\usepackage{nicefrac}       
\usepackage{microtype}      
\usepackage{xcolor}         

\usepackage{ragged2e}
\usepackage{tabularx}
\usepackage[pdftex]{graphicx}
\usepackage{makecell}
\usepackage{multirow}
\usepackage{booktabs}
\usepackage{adjustbox}
\usepackage{array}
\usepackage{wrapfig}
\usepackage{caption}
\usepackage{algorithm}
\usepackage{algpseudocode}
\usepackage{amsmath}
\usepackage{wrapfig}
\usepackage{placeins}
\usepackage[table]{xcolor}
\definecolor{lightskyblue}{rgb}{0.53, 0.81, 0.98}

\title{Rotation-Aligned Key Channel Pruning for Efficient Vision-Language Model Inference}

\author{%
Beomseok Kang, Dongwon Jo, Jiwon Song, Donghwee Son, Jae-Joon Kim \\
  Seoul National University\\
  \texttt{\{beomseok,kimjaejoon\}@snu.ac.kr} \\
}

\begin{document}

\maketitle

\begin{abstract}
Vision-Language Models suffer severe KV cache pressure at inference, as a single image often encodes into thousands of tokens. Most existing methods exploit token sparsity through token pruning, but permanently discarding visual content causes substantial degradation on fine-grained perception tasks. This motivates a complementary axis, feature sparsity: under a fixed KV cache budget, compressing the channel dimension preserves more visual tokens at the same memory cost. Prior Key channel pruning methods, however, face a structural trade-off: token-wise channel pruning is expressive but unstructured and slow, while head-wise approach is hardware-friendly but less robust. We resolve this with RotateK, a rotation-based structured Key channel pruning framework. RotateK applies an online PCA-based rotation that aligns token-dependent channel importance into a shared low-dimensional subspace, enabling accurate pruning under lightweight head-wise masks; a fused Triton attention kernel operates directly on sparse-channel Keys for efficient decoding. Experiments on two representative VLM backbones show that RotateK consistently outperforms prior Key channel pruning in both accuracy and decoding latency, while joint token-channel pruning improves over token-only baselines at matched KV cache budgets.

\end{abstract}

\section{Introduction}
Recent progress in Large Language Models (LLMs) has naturally extended to multi-modal systems such as Vision-Language Models (VLMs)~\cite{liu2024llavanext, chen2024internvl, bai2025qwen3}. By inheriting the broad world knowledge and reasoning capabilities of their language backbones, VLMs have demonstrated remarkable visual understanding across diverse tasks, from fine-grained perception~\cite{mathew2021docvqa, yu2016modeling} to complex reasoning over images~\cite{lu2023mathvista, lu2022learn} and videos~\cite{lin2024video, li2025videochat}. However, they inherit the inference bottlenecks of LLMs as well, most notably the KV cache, whose size grows linearly with sequence length. This issue is more severe in the visual setting, where a single image is encoded as hundreds to thousands of tokens~\cite{zhang2024long, shen2024longvu, chen2024longvila}, and video or multi-image inputs can easily reach tens of thousands~\cite{tu2024vl}. As a result, the visual KV cache often dominates memory usage during inference~\cite{xing2024pyramiddrop, chen2024image, wan2024look}, making its compression a key requirement for practical VLM deployment.

\begin{figure}[t]
	\begin{center}
	\includegraphics[width=0.95\linewidth]{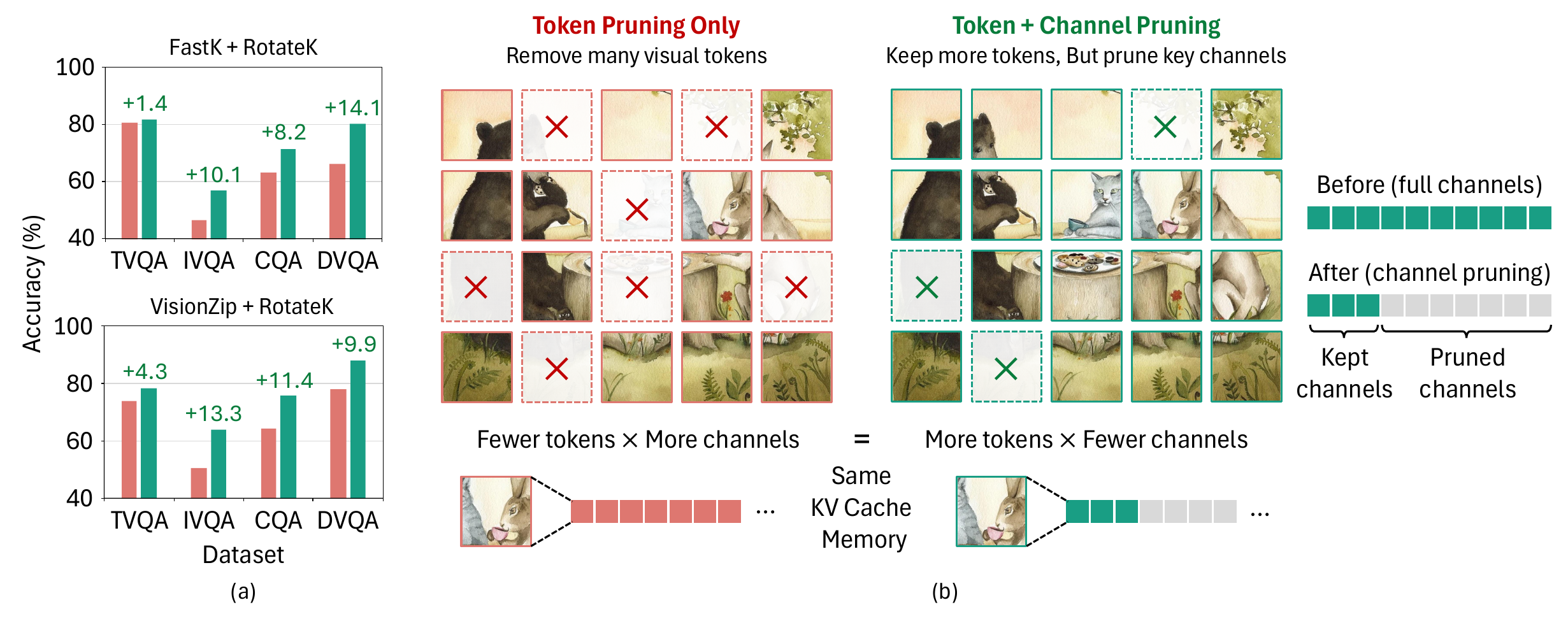}
    \end{center}
    \vspace{-0.4cm}
    \caption{\textbf{Token Pruning only vs. Token and Channel Pruning}. (a) Comparison between token pruning only (red) and joint token-channel pruning with RotateK (green) under similar KV cache budgets. For FastV, we compare 0.20 token sparsity against FastV (0.30$\times$ token) + RotateK (0.25$\times$ channel). For VisionZip, we compare 0.22 token sparsity against VisionZip (0.35$\times$ token) + RotateK (0.25$\times$ channel). (b) Unlike token pruning, which removes many visual tokens entirely, our approach preserves more visual tokens by additionally pruning less informative key channels under the same KV cache memory budget.}
    \label{fig:introduction}
    \vspace{-0.4cm}
\end{figure}

In recent years, visual KV cache compression has been predominantly driven by \textit{token pruning}~\cite{yang2025visionzip, chen2024image, alvar2025divprune, khaki2025sparsevila, ye2025fit}, which permanently discards visual tokens deemed less informative. While effective on many benchmarks, this strategy incurs irreversible information loss that degrades performance on tasks where relevant content is distributed broadly across the scene~\cite{alvar2025divprune}, such as document understanding~\cite{ masry2022chartqa, liu2024ocrbench} and visual grounding~\cite{wang2024cogvlm}. The KV cache, however, scales not only with the number of tokens $L$, but also with the channel dimension $d$. This suggests that, rather than relying solely on token sparsity, compression can also be achieved along the feature dimension. By incorporating \textit{feature sparsity} via pruning less informative channels, the same memory budget can be met while retaining more visual tokens, thereby mitigating the information loss inherent in token pruning (see Figure~\ref{fig:introduction}).

In practice, prior feature-dimension compression methods have primarily focused on pruning Key channels~\cite{xu2024think, liao2026spark, zhang2025leank}. Existing Key channel pruning methods largely fall into two regimes: head-wise pruning~\cite{xu2024think, zhang2025leank}, where a shared set of channels is pruned across tokens within each head, and token-wise pruning~\cite{liao2026spark}, where pruning decisions vary per token. These approaches reflect a trade-off between efficiency and expressivity: token-wise pruning achieves higher accuracy~\cite{liao2026spark}, suggesting that channel importance varies significantly across tokens, but requires a mask of size $\mathcal{O}(Ld)$ and incurs substantial memory and IO overhead during decoding; in contrast, head-wise pruning is far more efficient with a lightweight $\mathcal{O}(d)$ mask, but suffers from notable accuracy degradation at high compression ratios due to its inability to capture per-token differences. This raises a central question: Can token-dependent channel importance be captured within a structured, head-wise mask?

In this work, we propose \textbf{RotateK}, a rotation-based structured Key channel pruning framework for robust token–channel joint compression in VLMs. Our key insight is that token-dependent channel importance, while appearing highly unstructured in the original Key representation, becomes substantially more aligned under an appropriate orthogonal basis transformation. Based on this observation, RotateK computes an online PCA-based rotation that compresses visual Keys into a shared low-dimensional subspace across tokens, enabling efficient structured channel pruning while preserving more visual tokens under the same KV cache budget. As a result, RotateK substantially improves the robustness of KV cache compression compared to token-only pruning, while its structured sparsity patterns naturally simplify kernel design and facilitate efficient decoding acceleration. To the best of our knowledge, RotateK is the first dedicated study of Key channel pruning in VLM inference. Our key contributions are as follows:


\begin{itemize}
    \item We introduce token-channel joint compression for robust visual KV cache compression in VLMs. By additionally compressing the Key channel dimension, RotateK preserves more visual tokens under the same KV cache budget and substantially improves robustness over token-only pruning on fine-grained visual understanding tasks. 
    \item RotateK uses query-weighted PCA to concentrate query-relevant Key information into a shared low-dimensional subspace, minimizing perturbation to attention scores while enabling accurate structured channel pruning. This substantially narrows the accuracy gap between efficient head-wise pruning and expressive token-wise pruning.
    \item We develop a hardware-efficient implementation of RotateK by reducing online PCA overhead through Cholesky-based subspace iteration and enabling decoding acceleration via a custom Triton kernel over structured sparse channels. We further provide a detailed latency and memory analysis of Key channel pruning methods for long-context VLM inference.

\end{itemize}

\section{Motivational Study}


\begin{figure}[t]
	\begin{center}
	\includegraphics[width=\linewidth]{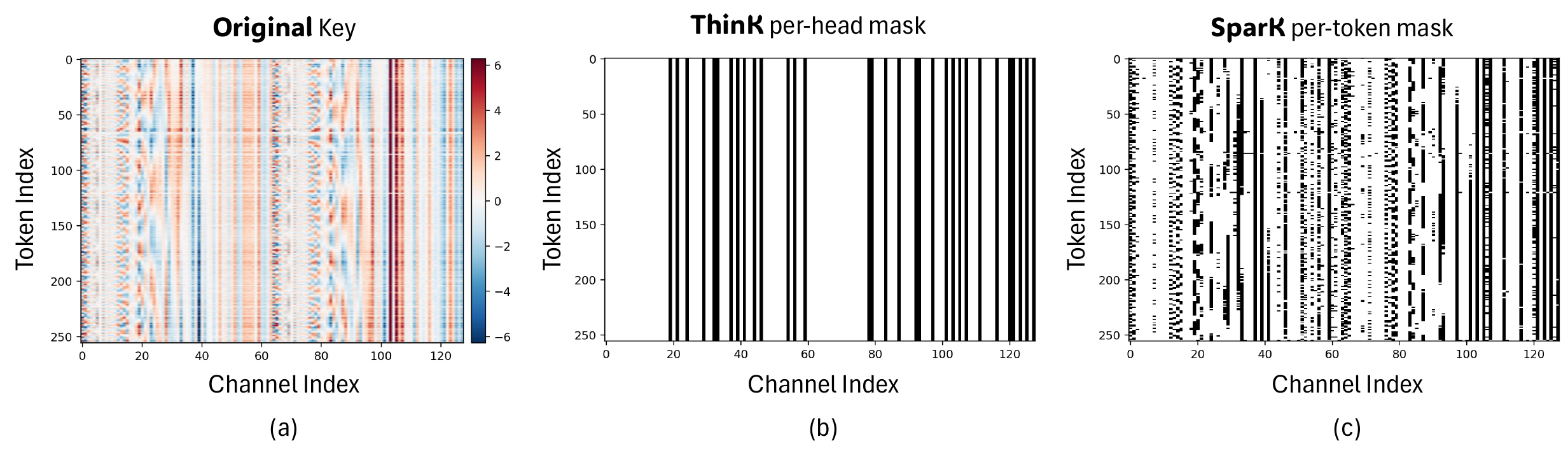}
    \end{center}
    \vspace{-0.3cm}
    \caption{\textbf{Key Channel Pruning Masks}. (a) Visualization of the original visual Key states, where few channels exhibit strong outlier patterns across tokens. (b) ThinK applies a shared head-wise mask, retaining the same channels for all tokens. (c) SparK applies token-wise masks that vary across tokens, reflecting token-dependent channel importance. The heterogeneous patterns suggest that informative channels are not aligned across tokens in the original channel basis. Results are experimented using LLaVA-NeXT-8B (layer-4 and head-4) on the TextVQA validation set.}
    \label{fig:motivation}
    \vspace{-0.3cm}
\end{figure}

\paragraph{Outliers in Visual Key Channels.} 
Key channel pruning is motivated by a structural property of Key states (see "Related Works" in Appendix~\ref{sup:related_works}): only a small subset of channels exhibit significantly larger magnitudes across tokens. Such channel-wise outliers have been widely observed in LLMs~\cite{liu2024kivi, xu2024think, zhang2025leank, liao2026spark, hooper2024kvquant}, and recent studies suggest that they primarily arise from the interaction between Rotary Position Embedding (RoPE) and learned Key projections~\cite{barbero2024round, zhang2025leank, qiao2025rethinking}. Since modern VLMs inherit the same transformer backbone and RoPE mechanism from their underlying LLMs, a similar outlier structure naturally emerges in visual Key states. Figure~\ref{fig:motivation}(a) visualizes the Key tensor, where channels near indices $40$ and $100$ appear as clear outliers.

\paragraph{Token-dependent Channel Importance.} However, such universally dominant channels are few in number, suggesting that preserving only these outliers may not be sufficient for robust Key channel pruning. To further investigate this, we compare the pruning masks retaining 32 channels ($25\%$) produced by ThinK (head-wise pruning)~\cite{xu2024think} and SparK (token-wise pruning)~\cite{liao2026spark} in Figure~\ref{fig:motivation}(b)-(c). Notably, SparK's mask contains only a small number of consistently preserved channels, appearing as the solid black columns in Figure~\ref{fig:motivation}(c). While these correspond to universally important outlier channels, most retained channels vary substantially across tokens, indicating that channel importance is largely token-dependent. This dependency becomes particularly pronounced under RoPE, where channels near indices $20$ and $80$ exhibit token-dependent magnitude oscillations induced by rotational encoding, making no fixed subset uniformly informative across tokens.

\paragraph{Toward Token-invariant Basis.} This token dependency creates a fundamental challenge for structured Key channel pruning, which relies on a static channel mask shared across tokens. Under RoPE, informative directions vary across token positions, making structured pruning in the standard channel basis inherently suboptimal. We interpret this phenomenon as a \emph{basis mismatch}: although each token admits a compact informative subspace, those directions are not aligned across tokens in the original channel coordinates. This suggests a natural remedy: instead of pruning channels directly in the original basis, we first seek a transformed coordinate system, realized through an orthogonal rotation of the Key space, where informative directions become more consistently aligned across tokens. In such a basis, a single structured mask can better preserve the informative subspace across diverse tokens, enabling more robust structured channel pruning at aggressive compression ratios.

\section{Proposed Methods}

\begin{figure}[t]
	\begin{center}
	\includegraphics[width=\linewidth]{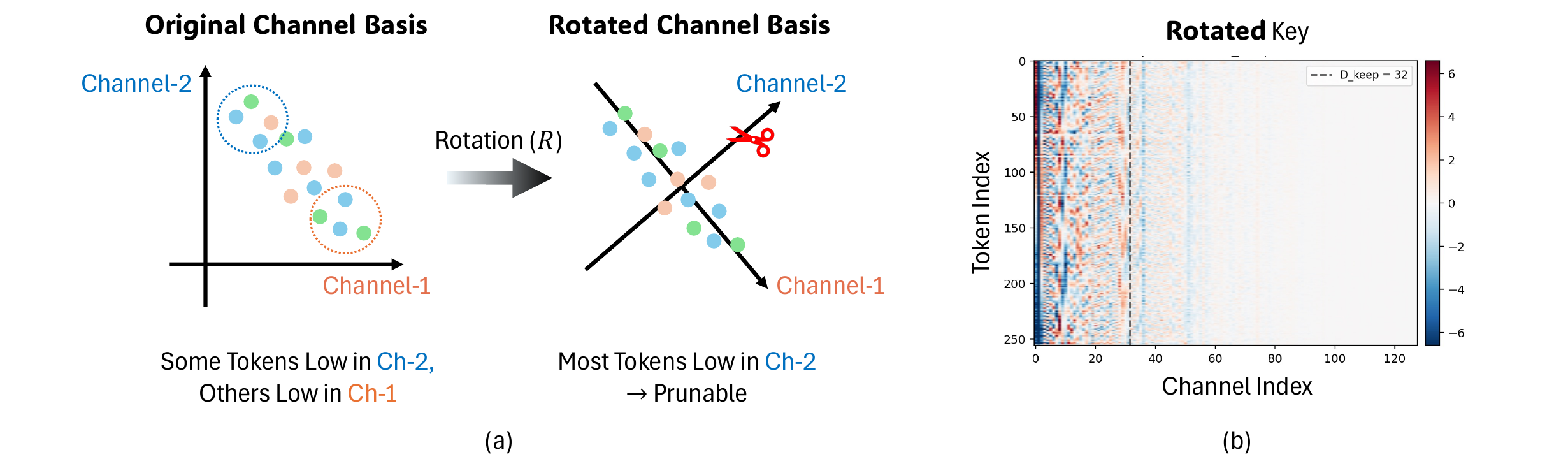}
    \end{center}
    \vspace{-0.3cm}
    \caption{\textbf{High-level Idea of RotateK}. (a) In the original channel basis, different visual tokens exhibit different low-importance channels, making structured head-wise pruning ineffective. RotateK applies an orthogonal rotation R to align token-dependent importance into a shared channel basis, where most tokens consistently exhibit low activations on the same channels. (b) Visualization of the rotated Key states after applying RotateK, where informative channels become concentrated into a smaller subset of aligned channels across tokens, enabling effective structured key channel pruning.}
    \label{fig:high_level_idea}
    \vspace{-0.3cm}
\end{figure}

\subsection{Background}

\paragraph{Rotation Matrix.}
To make channel-wise pruning more effective, we apply a rotation in the channel space. The key idea is to transform the channel basis so that token-wise importance is concentrated in a small subset of channels, making the less informative ones easier to discard (see Figure~\ref{fig:high_level_idea}(a)).

A rotation matrix $R \in \mathbb{R}^{d \times d}$, where $d$ denotes the channel dimension, is an orthogonal matrix satisfying $RR^{\top} = I$ (and typically $\det(R) = 1$). It preserves inner products, and hence the norms and pairwise angles of vectors. Multiplying features by $R$ thus amounts to a change of orthonormal basis: no information is lost, but information is redistributed across channels. Crucially, the nature of this redistribution depends entirely on the choice of $R$. Some rotations, such as Hadamard transforms, deliberately spread energy uniformly across channels to suppress activation outliers; in contrast, we seek rotations that do the opposite, \textit{i.e.}, concentrating information into a small subset of channels so that the remaining channels can be safely pruned.

\paragraph{Variance Concentration via PCA.}
Our approach is closely related to principal component analysis (PCA), which concentrates the variance of the data along a few rotated axes, or equivalently, identifies a low-rank subspace that well approximates the Key states. Such a subspace can be obtained via eigendecomposition of the covariance matrix. However, because the optimal subspace varies across visual inputs, performing an eigendecomposition for every input at inference time would substantially inflate latency. It is therefore essential to construct the rotation matrix $R$ in a way that is efficient enough to be generated on the fly.

\subsection{Overview of RotateK}
\label{sec:overview}

Our goal is to compress the Key states $K \in \mathbb{R}^{N \times d}$ of
$N$ visual tokens (observed at prefill) from $d$ to $k < d$
channels with minimal perturbation to the attention scores
$q_t K^{\top}$ at decode step $t$, where $q_t \in \mathbb{R}^{d}$ is the
new query. RotateK achieves this in two steps:
\textbf{(1) rotate} the channel space with an orthogonal matrix
$R \in \mathbb{R}^{d \times d}$, and \textbf{(2) prune} the less
informative channels in the rotated basis. Figure~\ref{fig:inference_flow} illustrates the overall inference flow of RotateK (see algorithmic overview in Appendix~\ref{sup:algorithm}).

\paragraph{Step (1): Rotation is lossless.}
RotateK computes a rotation matrix $R$ ($R^{\top} R = I_d$) \emph{once}
at the end of prefill, separately for each attention head, and reuses it
for every subsequent decode step; the construction of $R$ is described
in the following paragraphs. If we rotate both keys and queries by $R$ and
retain all $d$ channels, the attention scores are exactly preserved,
\begin{equation}
    (q_t R)\, (K R)^{\top}
    \;=\; q_t\, R R^{\top}\, K^{\top}
    \;=\; q_t K^{\top},
    \label{eq:rotation-lossless}
\end{equation}
since $R R^{\top} = I_d$. Step (1) alone therefore introduces no
approximation error and no compression. It merely re-expresses the
channels in a different basis.

\begin{figure}[t]
	\begin{center}
	\includegraphics[width=\linewidth]{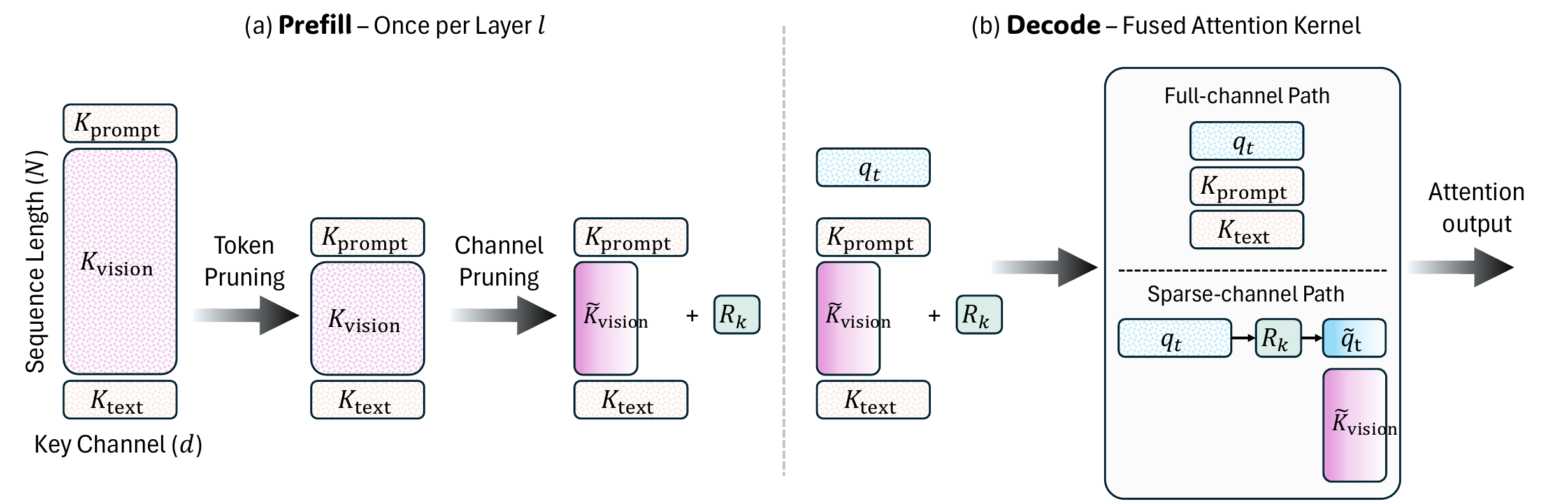}
    \end{center}
    \vspace{-0.3cm}
    \caption{\textbf{Inference Flow of RotateK}. (a) Following visual token compression, RotateK applies head-wise rotation-based channel pruning to visual Key states during prefill, caching the compressed KV states together with the corresponding rotation matrices. (b) During decoding, queries are transformed using the cached rotations, while a fused attention kernel (see Appendix~\ref{sup:rotatek-kernel} for details) combines full-channel attention for prompt/text tokens with sparse-channel attention for visual tokens.}
    \label{fig:inference_flow}
    \vspace{-0.3cm}
\end{figure}

\paragraph{Step (2): Channel pruning introduces the approximation.}
Compression appears once we prune channels \emph{in the rotated basis}.
Let $R_k \in \mathbb{R}^{d \times k}$ collect the top-$k$ columns of $R$,
so that $R_k^{\top} R_k = I_k$. At prefill, RotateK caches the truncated
rotated keys $\tilde K = K R_k \in \mathbb{R}^{N \times k}$ in place of
the full $N \times d$ matrix $K$; at decode, the incoming query is
rotated to $\tilde q_t = q_t R_k \in \mathbb{R}^{k}$ and computed with $\tilde K$. Notably, pruning the rotated channel is essentially structured (\textit{i.e.}, head-wise) as a rotation matrix is shared across tokens. The resulting attention scores no longer match
$q_t K^{\top}$:
\begin{equation}
    \tilde q_t \tilde K^{\top}
    \;=\; (q_t R_k)\,(K R_k)^{\top}
    \;=\; q_t\, P_k\, K^{\top},
    \qquad P_k := R_k R_k^{\top},
    \label{eq:rotated-attn}
\end{equation}
because $P_k$ is a rank-$k$ projector and $P_k \neq I_d$ whenever
$k < d$. The approximation error introduced by step (2) is therefore
exactly:
\begin{equation}
    q_t K^{\top} - \tilde q_t \tilde K^{\top}
    \;=\; q_t\, (I_d - P_k)\, K^{\top},
    \label{eq:approx-error}
\end{equation}
the projection of $q_t K^{\top}$ onto the $d - k$ discarded channels of
the rotated basis. The remainder of this section is concerned with choosing $R$, and hence $R_k$ and $P_k$, so that this residual is small for the decode queries $q_t$ that will arrive after $R_k$ is
fixed.


\paragraph{Query-Weighted PCA.} Our goal is to rotate the Key channels such that the information relevant to the attention scores is concentrated into a smaller set of leading channels, making subsequent channel pruning less destructive. A natural approach is to apply PCA directly to the Key states $K$, which preserves directions with large Key variance. However, the pruning error is ultimately measured on the attention scores $q_t K^\top$, not on the Keys themselves. Consequently, channels where both the Key activations and query magnitudes are large should be preserved preferentially. 

We capture this intuition by weighting each Key channel according to the
typical magnitude of recent queries. Let $Q_W \in \mathbb{R}^{W \times d}$ denote the window of the $W$ most recent prefill queries (we use $W=32$, following~\cite{xu2024think}), and let $\boldsymbol{\sigma}_W \in \mathbb{R}^{d}$ collect their per-channel $\ell_2$ norms: $(\boldsymbol{\sigma}_W)_j = \|(Q_W)_{:,j}\|_2$.

We then apply PCA to the mean-centered, query-weighted Key activations $K_q = (K - \mu)\,\mathrm{diag}(\boldsymbol{\sigma}_W)$, where $\mu \in \mathbb{R}^d$ is the per-head channel mean of $K$ over the $N$ visual tokens. The rotation matrix $R = [r_1, \dots, r_d]$ collects the eigenvectors of the query-weighted covariance given by:
\begin{equation}
    C_q
    \;=\;
    K_q^{\top} K_q
    \;=\;
    \mathrm{diag}(\boldsymbol{\sigma}_W)\,
    C\,
    \mathrm{diag}(\boldsymbol{\sigma}_W),
\end{equation}
where $C = (K - \mu)^{\top} (K - \mu)$ is the centered Key covariance. 

We obtain the rotation matrix $R=[r_{1}, r_{2}, ... ,  r_{d}]$ by eigendecomposing $C_{q}$, where the columns of $R$ are eigenvectors of $C_{q}$ ordered by decreasing eigenvalue magnitude. The first $k$ columns are retained as $R_k \in \mathbb{R}^{d \times k}$ for channel pruning. The mean $\mu$ is not in general aligned with the column space of $R_k$, so its discarded component reintroduces a constant bias $q_t^{\top} (I_d - R_k R_k^{\top})\,\mu$ in the rotated attention, which is added back during decoding. Comparison between query-weighted vs. query-agnostic PCA is available in Appendix~\ref{sup:rotatek-ablation}.

\subsection{Hardware-efficient Implementation}

\paragraph{Post-Hoc Reweighting.}
A naive implementation of query-weighted PCA would explicitly construct the rescaled Key matrix $K_q \in \mathbb{R}^{N \times d}$, introducing an additional intermediate of the same size as $K$. This increases memory traffic by materializing and accessing another $N \times d$ tensor, and incurs an extra $\mathcal{O}(Nd)$ cost for channel-wise rescaling.

Instead, we observe that the query-weighted covariance can be formed directly from the original covariance $C = K^\top K$:
\begin{equation}
    C_q
    =
    K_q^\top K_q
    =
    \mathrm{diag}(\boldsymbol{\sigma}_W)
    \, C \,
    \mathrm{diag}(\boldsymbol{\sigma}_W)
    =
    (\boldsymbol{\sigma}_W \boldsymbol{\sigma}_W^\top)
    \odot
    C,
\end{equation}
where $\odot$ denotes the Hadamard (element-wise) product. Thus, once $C$ is computed, query-weighting requires only a rank-one outer product and an element-wise multiplication on a $d \times d$ matrix, resulting in an additional complexity of $\mathcal{O}(d^2)$ per head (typically $d \ll N$). Crucially, this overhead is independent of the context length $N$, making the query-weighting step effectively free in the long-context regime targeted by KV-cache pruning.



\paragraph{Cholesky-based Subspace Iteration.}
While the additional overhead of query-weighted PCA itself is marginal, PCA still introduces noticeable prefill latency. The standard pipeline computes the covariance at $\mathcal{O}(N d^2)$ followed by a full eigendecomposition at $\mathcal{O}(d^3)$. Although this is theoretically small compared to the attention cost $\mathcal{O}(N^2 d)$ ($N \gg d$), wall-clock latency is dominated less by arithmetic throughput than by GPU kernel-dispatch overhead. In particular, functions such as \texttt{torch.linalg.eigh} internally launch \emph{tens of} fragmented CUDA kernels for tridiagonalisation, the inner eigensolver, and eigenvector reconstruction.

To reduce this overhead, RotateK replaces full eigendecomposition with a Cholesky-based subspace iteration that directly estimates only the top-$k$ eigenspace. Starting from a random basis $V^{(0)} \in \mathbb{R}^{d \times k}$, each iteration applies $V^{(t)} \leftarrow C_q V^{(t-1)}$, progressively aligning $V$ with the dominant eigenspace of $C_q$, followed by a Cholesky-QR orthonormalisation to preserve the rank-$k$ structure (see Appendix~\ref{sup:cholesky_iteration}); in practice, $T{=}5$ iterations already match the accuracy of full eigendecomposition (see Appendix~\ref{sup:rotatek-ablation}). Per iteration costs $\mathcal{O}(d^2 k)$, dominated by the GEMM $C_q V$, and is comparable
to $\mathcal{O}(d^3)$ in flops at our matrix sizes--but issues only \emph{four} CUDA kernels per iteration (two GEMMs, one $k \times k$ Cholesky, one triangular solve), with fixed tensor shapes that allow CUDA-graph capture of the entire $T$-step loop and an order-of-magnitude lower dispatch latency in practice.

\section{Experimental Results}


\paragraph{Dataset and Implementation Details.} Following the \texttt{lmms-eval}~\cite{zhang2025lmms} protocol, we evaluate on VQA benchmarks (TextVQA, InfoVQA, DocVQA, ChartQA, VizWiz-VQA) and open-ended generation benchmarks (LLaVA-in-the-Wild, MM-Vet). Throughout the paper, TextVQA, InfoVQA, DocVQA, and ChartQA are abbreviated as TVQA, IVQA, DVQA, and CQA, respectively. For latency evaluation, we compare Triton-based FlashAttention kernels against our Triton-based sparse attention kernel under identical settings. Unless otherwise specified, RotateK operates in an online calibration-free mode, recomputing the PCA-based rotation matrix during each prefill stage. All experiments are conducted on a single NVIDIA A100-80GB GPU.

\paragraph{Baselines.} We evaluate RotateK on LLaVA-NeXT-8B (\texttt{llama3-llava-next-8b-hf}) and Qwen2.5-VL-7B-Instruct. To isolate the effect of channel pruning, RotateK is integrated into two orthogonal token-pruning frameworks: VisionZip~\cite{yang2025visionzip} and FastV~\cite{chen2024image}. We compare against two recent training-free Key channel pruning techniques, ThinK~\cite{xu2024think} and SparK~\cite{liao2026spark} under identical memory budgets, and additionally report token-only and channel-only baselines. Detailed hyperparameter settings are provided in Appendix~\ref{sup:experimental_deatils}.

\begin{figure}[t]
	\begin{center}
	\includegraphics[width=\linewidth]{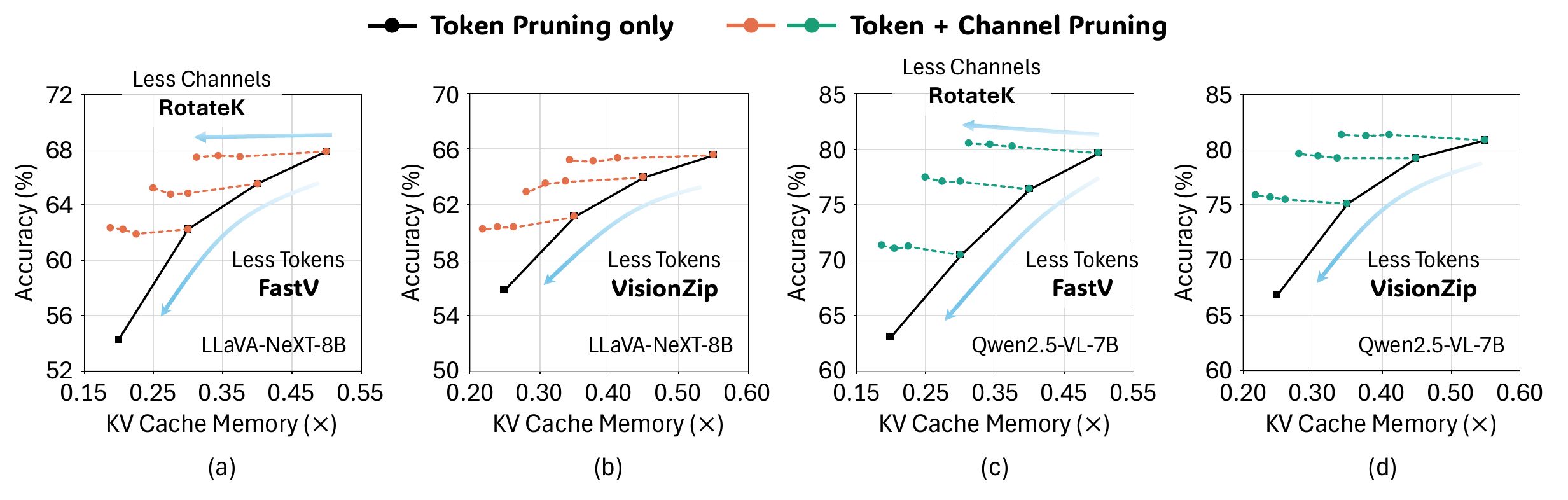}
    \end{center}
    \vspace{-0.4cm}
    \caption{\textbf{Accuracy--KV Cache Trade-offs}. Accuracy--memory trade-offs on ChartQA using LLaVA-NeXT-8B (orange) and Qwen2.5-VL-7B (green) under varying token and channel sparsity ratios. Compared to token-only pruning (black), RotateK jointly prunes tokens and Key channels, consistently achieving higher accuracy under the same KV cache budget.}
    \label{fig:token_vs_channel}
    \vspace{-0.2cm}
\end{figure}

\begin{table*}[t]
	\renewcommand{\arraystretch}{1.05}
	\setlength{\tabcolsep}{4pt}
	\fontsize{8.5pt}{8.5pt}\selectfont
	\caption{\textbf{Comparison under Fixed KV Cache Budgets}. Accuracy comparison between token-only pruning and joint token-channel pruning under identical KV cache budgets. Across both LLaVA-NeXT-8B and Qwen2.5-VL-7B-Instruct, RotateK consistently improves over token-only pruning and prior Key channel pruning. More results are provided in Table~\ref{tab:appendix_additional_comparison} (Appendix).}
    \vspace{-0.2cm}    
	\label{tab:main_comparison}
	\begin{center}
	\begin{tabular}{l| ccc | ccccc | c}

		\toprule

		\textbf{Method} & \textbf{Token} & \textbf{Channel} & \textbf{KV Cache} & \textbf{TVQA} & \textbf{IVQA} & \textbf{CQA} & \textbf{DVQA} & \textbf{VizWiz} & \textbf{mean}\\

		\midrule
        \multicolumn{9}{c}{\textsc{Llama3-LLaVA-NeXT-8B}} \\
        \midrule
        Baseline & 1.00$\times$ & 1.00$\times$ & 1.00$\times$ & 65.40 & 32.27 & 69.08 & 72.44 & 57.50 & 59.34 \\
        \midrule

        FastV$_\text{token only}$ & 0.25$\times$ & 1.00$\times$ & 0.25$\times$ & 62.69 & 25.91 & 59.64 & 56.39 & 58.56 & 52.64\\
        FastV + ThinK & 0.40$\times$ & 0.25$\times$ & 0.25$\times$ & 54.07 & 25.51 & 60.56 & 51.72 & 57.70 & 49.91\\
        \rowcolor{lightskyblue!30} FastV + RotateK & 0.40$\times$ & 0.25$\times$ & 0.25$\times$ & 62.67 & 28.37 & 65.25 & 62.46 & 58.34 & 55.42 \\

        
        \midrule
        VisionZip$_\text{token only}$ & 0.28$\times$ & 1.00$\times$ & 0.28$\times$ & 63.33 & 27.70 & 57.32 & 64.08 & 58.97 & 54.28 \\
        VisionZip + ThinK & 0.45$\times$ & 0.25$\times$ & 0.28$\times$ & 54.22 & 27.81 & 58.52 & 53.77 & 57.81 & 50.43 \\
        \rowcolor{lightskyblue!30} VisionZip + RotateK & 0.45$\times$ & 0.25$\times$ & 0.28$\times$ & 62.62 & 28.97 & 62.88 & 66.99 & 58.94 & 56.08 \\


        \midrule
        \multicolumn{9}{c}{\textsc{Qwen2.5-VL-7B-Instruct}} \\
        \midrule

        Baseline & 1.00$\times$ & 1.00$\times$ & 1.00$\times$ & 82.92 & 80.12 & 83.04 & 94.40 & 70.58 & 82.21 \\


        \midrule
        FastV$_\text{token only}$ & 0.25$\times$ & 1.00$\times$ & 0.25$\times$ & 81.49 & 52.76 & 67.44 & 75.04 & 69.13 & 69.17 \\
        FastV + ThinK & 0.40$\times$ & 0.25$\times$ & 0.25$\times$ & 77.81 & 63.01 & 77.12 & 81.74 & 69.25 & 73.79 \\
        \rowcolor{lightskyblue!30} FastV + RotateK & 0.40$\times$ & 0.25$\times$ & 0.25$\times$ & 82.38 & 65.79 & 77.40 & 87.28 & 70.32 & 76.63 \\   
 
        
        \midrule
        VisionZip$_\text{token only}$ & 0.28$\times$ & 1.00$\times$ & 0.28$\times$ & 76.84 & 58.46 & 70.32 & 84.71 & 68.97 & 71.86 \\
        VisionZip + ThinK & 0.45$\times$ & 0.25$\times$ & 0.28$\times$ & 75.72 & 67.13 & 79.64 & 85.83 & 68.93 & 75.45 \\
        \rowcolor{lightskyblue!30} VisionZip + RotateK & 0.45$\times$ & 0.25$\times$ & 0.28$\times$ & 80.84 & 70.51 & 79.56 & 91.29 & 70.13 & 78.47 \\


        \bottomrule
	\end{tabular}
	\end{center}
    \vspace{-0.4cm}    
\end{table*}

\subsection{Comparison with Token Pruning}

\paragraph{Accuracy--KV Cache Trade-offs.} To further investigate the accuracy-memory trade-offs of token pruning versus joint token-channel pruning, Figure~\ref{fig:token_vs_channel} compares their performance on ChartQA under varying compression ratios, \textit{e.g.}, channel sparsity ratios of $0.500$, $0.375$, and $0.250$ (less channels) are evaluated. The horizontal points labeled ``Less Channels'' apply additional channel pruning on top of fixed token sparsity ratios. While token-only pruning degrades rapidly at aggressive compression ratios, RotateK consistently achieves higher accuracy under the same KV cache budget, remaining robust even at $75\%$ channel sparsity. These results demonstrate that channel pruning complements token pruning, enabling more robust KV cache compression in VLMs.

\begin{figure}[t]
	\begin{center}
	\includegraphics[width=\linewidth]{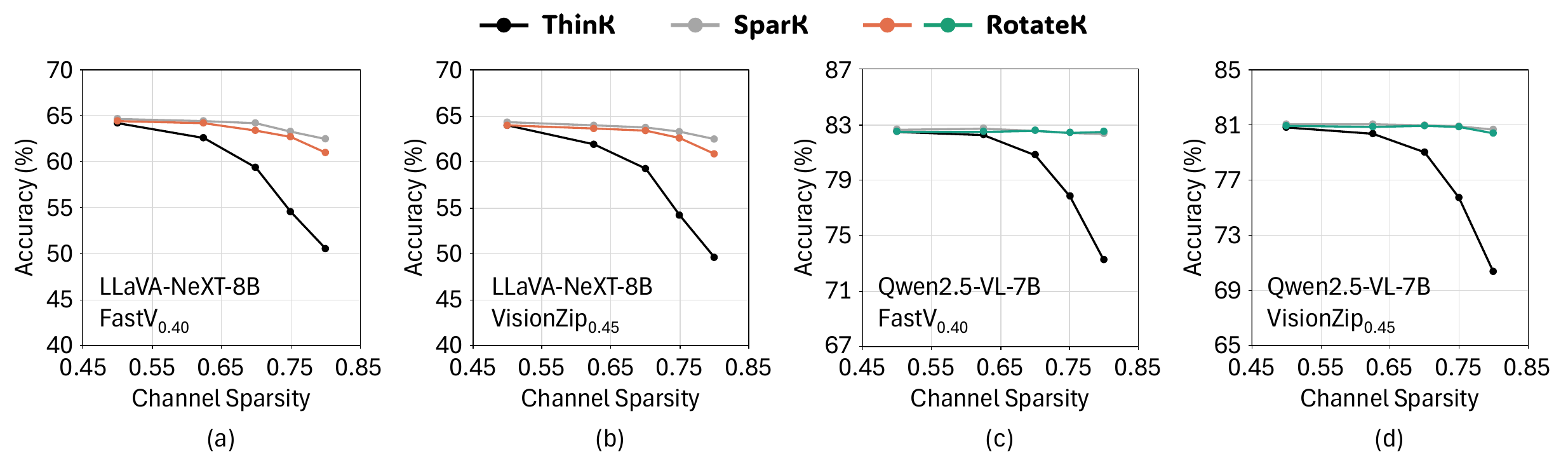}
    \end{center}
    \vspace{-0.3cm}
    \caption{\textbf{Channel Pruning Robustness.} Accuracy under increasing channel sparsity ratios for ThinK, SparK, and RotateK integrated with FastV and VisionZip on LLaVA-NeXT-8B and Qwen2.5-VL-7B. While ThinK rapidly degrades at high sparsity ratios, RotateK consistently maintains strong performance even at aggressive channel pruning levels, demonstrating improved robustness of structured Key channel pruning.}
    \label{fig:think_spark}
    \vspace{-0.3cm}
\end{figure}

\paragraph{Comparison under Fixed KV Cache Budgets.} Table~\ref{tab:main_comparison} compares RotateK across five benchmarks under two KV cache memory budgets. We observe that the degradation caused by token-only pruning varies substantially across datasets, even at similar token sparsity ratios. In particular, InfoVQA on Qwen2.5-VL-7B suffers severe degradation under token-only pruning, where RotateK consistently improves accuracy by more than $10\%$.

RotateK also consistently outperforms ThinK across most settings, with especially large gains on DocVQA for LLaVA-NeXT-8B. Across 40 evaluated cases, ThinK surpasses RotateK in only five cases, and only by marginal differences below $0.6\%$. Notably, the benefits of RotateK are not limited to scenarios where token-only pruning collapses; even on relatively robust benchmarks such as TextVQA and VizWiz, RotateK consistently maintains comparable or better performance under the same KV cache budget.

\paragraph{Comparison on Open-Ended Benchmarks.} As, channel pruning methods typically do not impact the prefill and rather impact the decode, we compare how robust token-channel pruning is on longer-generation tasks, beyond short generation VQA tasks. Table ~\ref{tab:open_ended} presents the accuracy in two open-ended benchmarks, LLava-in-the-wild and MMVet, where typically tens of tokens are generated to provide the details in the scene. We consistently observe Token Pruning + RotateK $>$ Token Pruning + ThinK $\approx$ Token Pruning only. This further confirms the generality of our strategy to jointly combine token pruning and RotateK.

\begin{wraptable}{r}{0.50\linewidth}  
\renewcommand{\arraystretch}{1.0}
\setlength{\tabcolsep}{4.5pt}
\fontsize{8.5pt}{8.5pt}\selectfont
\vspace{-35pt}  
\centering

\caption{\textbf{Comparison on Open-Ended Benchmarks}. Results on LLaVA-in-the-Wild and MM-Vet using LLaVA-NeXT-8B and Qwen2.5-VL-7B-Instruct. RotateK consistently improves over token-only pruning and prior Key channel pruning methods under the same KV cache budget. Subscripts denote token and channel sparsity ratios.}

\begin{tabular}{l|cc}
        
\toprule
\textbf{Method} & \textbf{Llava-Wild} & \textbf{MMVet} \\

\midrule
\multicolumn{3}{c}{\textsc{Llama3-LLaVA-NeXT-8B}} \\
\midrule

Baseline & 84.10 & 27.14 \\
\midrule
FastV$_{0.1875}$ & 84.60 & 24.50 \\ 
FastV$_{0.30}$ + ThinK$_{0.25}$ & 80.40 & 25.05 \\
\rowcolor{lightskyblue!30} FastV$_{0.30}$ + RotateK$_{0.25}$  & 85.50 & 26.97 \\

\midrule

VisionZip$_{0.22}$ & 81.10 & 24.59 \\ 
VisionZip$_{0.35}$ + ThinK$_{0.25}$ & 75.30 & 23.72 \\
\rowcolor{lightskyblue!30} VisionZip$_{0.35}$ + RotateK$_{0.25}$ & 84.70 & 26.97 \\

\midrule
\multicolumn{3}{c}{\textsc{Qwen2.5-VL-7B-Instruct}} \\
\midrule

Baseline & 109.30 & 45.55 \\
\midrule
FastV$_{0.1875}$ & 100.00 & 32.61 \\ 
FastV$_{0.30}$ + ThinK$_{0.25}$ & 99.80 & 32.61 \\
\rowcolor{lightskyblue!30} FastV$_{0.30}$ + RotateK$_{0.25}$ & 106.90 & 41.24 \\

\midrule

VisionZip$_{0.22}$ & 104.00 & 34.77 \\ 
VisionZip$_{0.35}$ + ThinK$_{0.25}$ & 99.10 & 35.00 \\
\rowcolor{lightskyblue!30} VisionZip$_{0.35}$ + RotateK$_{0.25}$ & 106.20 & 44.04 \\
 
\midrule

\end{tabular}
\label{tab:open_ended}
\vspace{-10pt}  
\end{wraptable}
\subsection{Comparison with Channel Pruning}

\paragraph{Structured vs. Unstructured Channel Pruning.} Since RotateK aims to concentrate token-dependent channel importance into a smaller subset of channels, we examine how effectively it closes the performance gap between head-wise (structured; \textit{e.g.}, ThinK) and token-wise (unstructured; \textit{e.g.}, SparK) channel pruning. Figure~\ref{fig:think_spark} shows the accuracy-channel sparsity trade-offs on TextVQA across two VLMs and two token pruning methods. At 50\% channel sparsity, all methods achieve similar accuracy. However, as sparsity increases, ThinK shows clear degradation, whereas RotateK maintains accuracy comparable to SparK across most sparsity levels. SparK achieves slightly higher accuracy at 80\% sparsity in LLaVA, likely due to its mean-based interpolation for pruned channels.

\begin{figure}[t]
	\begin{center}
	\includegraphics[width=\linewidth]{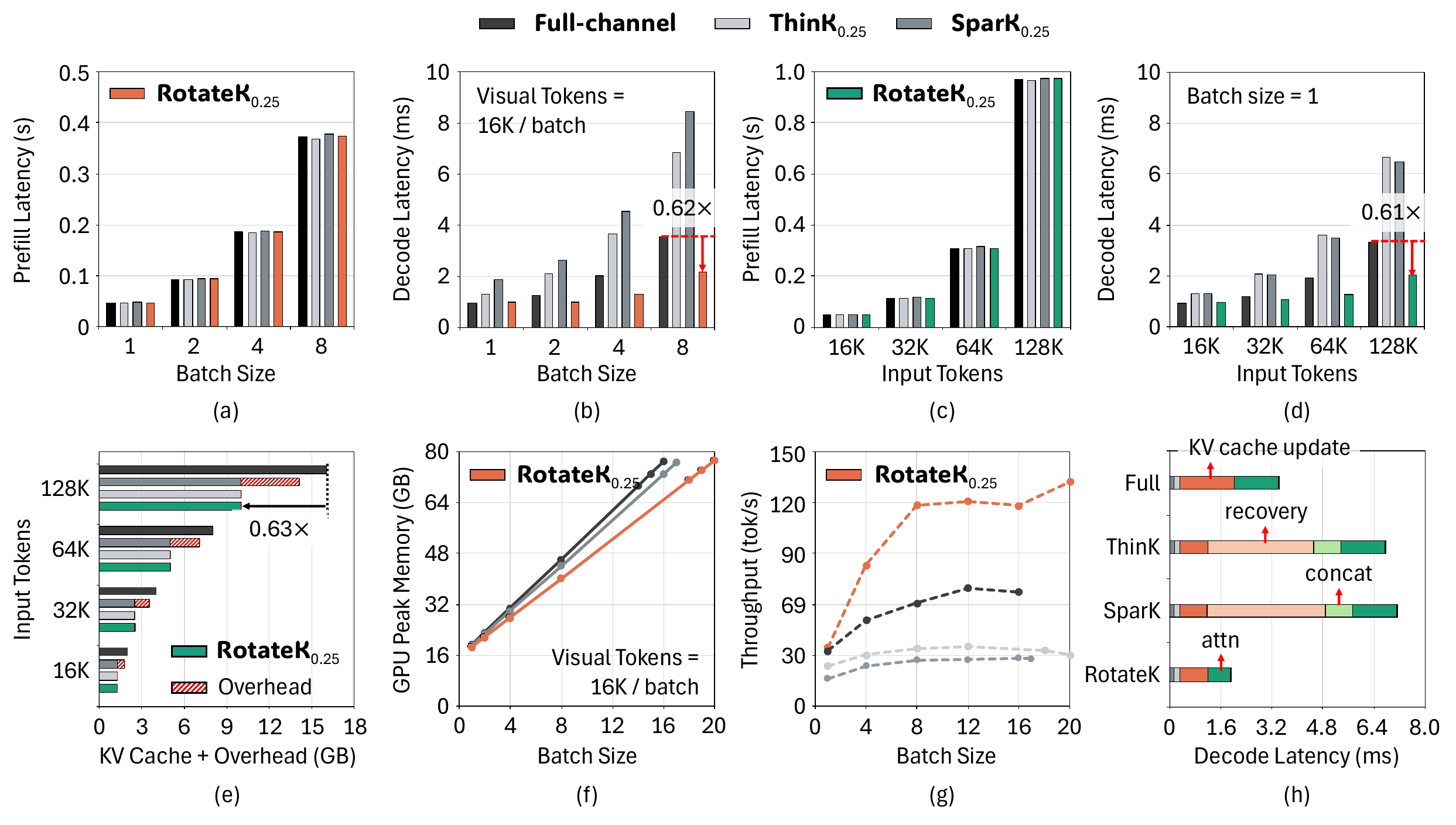}
    \end{center}
    \vspace{-0.3cm}
    \caption{\textbf{Latency and Memory Analysis.} Comparison of RotateK, ThinK, and SparK on prefill latency, decoding latency, KV cache memory, GPU peak memory, and generation throughput. (a,c) RotateK introduces negligible prefill overhead across varying batch sizes and sequence lengths. (b,d) Unlike ThinK and SparK, which reconstruct full-channel Keys before attention, RotateK directly performs sparse-channel attention and substantially reduces decoding latency. (e) RotateK incurs significantly smaller memory overhead than prior methods, leading to lower effective KV cache usage. (f,g) The reduced memory footprint enables larger batch sizes and higher generation throughput. (h) Decode latency breakdown, showing that full-channel Key recovery dominates the latency of prior channel pruning methods.}
    \label{fig:efficiency}
    \vspace{-0.3cm}
\end{figure}

\subsection{Computational Efficiency}
While prior channel pruning methods such as ThinK and SparK analyze accuracy and memory overhead, their latency behavior remains underexplored. We therefore provide a detailed analysis of the latency and memory overhead of key channel pruning methods.

\paragraph{Prefill Latency.} ThinK, SparK, and RotateK all introduce additional prefill computation to identify informative channels. Although RotateK performs online PCA with $\mathcal{O}(Nd^2)$ complexity, Figure~\ref{fig:efficiency}(a,c) shows that all three methods achieve prefill latency comparable to the full-channel baseline in practice.

\paragraph{Decode Latency.} 
Decoding latency exhibits a markedly different trend (Figure~\ref{fig:efficiency}(b,d)). As input length or batch size increases, ThinK and SparK become progressively slower, eventually reaching up to $1.8\times$ higher latency than the full-channel baseline. Although these methods reduce KV cache footprint through channel pruning, they reconstruct dense full-channel Keys during decoding, concatenate them with unpruned Keys (\textit{e.g.}, recent tokens), and apply the standard \texttt{FlashAttention2} kernel on the recovered tensor. This reconstruction introduces substantial memory IO overhead, largely negating the benefit of sparse-channel KV caches.

In contrast, RotateK employs a custom Triton-based attention kernel that directly operates on sparse-channel Keys without reconstruction. Its structured sparsity further enables a fused attention design over both sparse- and full-channel Keys, allowing sparse KV caches to translate into real decoding speedups. Figure~\ref{fig:efficiency}(h) further confirms that recovery and concatenation dominate the additional latency of ThinK and SparK, whereas RotateK eliminates these overheads entirely.



\paragraph{Memory Overhead.} Figure~\ref{fig:efficiency}(e,f) compares KV cache size and GPU peak memory. Notably, SparK introduces substantial memory overhead due to its token-wise pruning masks, largely offsetting the benefit of channel pruning. In contrast, ThinK and RotateK use head-wise sparsity and incur negligible overhead, enabling larger feasible batch sizes. However, larger batch sizes do not necessarily translate into higher throughput. As shown in Figure~\ref{fig:efficiency}(g), ThinK can still underperform the full-channel baseline due to its degraded decoding latency, whereas RotateK achieves consistently higher throughput.
\section{Conclusion}

We presented RotateK, a rotation-based structured Key channel pruning framework for efficient VLM inference. RotateK aligns token-dependent channel importance through an online PCA-based rotation, enabling accurate structured pruning with lightweight head-wise masks. By jointly compressing visual tokens and Key channels, RotateK preserves more visual information under the same KV cache budget and consistently improves robustness over token-only pruning on fine-grained visual understanding tasks. Furthermore, our hardware-efficient implementation, including Cholesky-based subspace iteration and a fused Triton sparse attention kernel, translates sparse-channel KV caches into actual decoding speedups with negligible memory overhead.

\def\bibfont{\small}

\bibliographystyle{unsrt} 
\medskip

\newpage
\bibliography{references}

\newpage
\appendix

\clearpage
\section{Related Works}
\label{sup:related_works}

\subsection{Visual Token Pruning}
Token-axis compression of the visual KV cache falls into two regimes. Pre-LLM
methods, exemplified by VisionZip~\cite{yang2025visionzip} and
LLaVA-PruMerge~\cite{shang2025llava}, discard tokens at the vision encoder
using saliency signals such as [CLS] attention~\cite{zhang2025beyond} or token-flow
propagation~\cite{tong2025flowcut}; lacking access to the language query, they
risk removing tokens critical to the downstream task. In-LLM methods defer
pruning until after vision-text attention: FastV~\cite{chen2024image} drops
visual tokens with low post-prefill attention, with extensions adding adaptive
layer ratios~\cite{ye2025fit}, token recycling~\cite{zhang2024sparsevlm},
or dual attention filtering~\cite{liu2024multi}. Yet visual token pruning
still incurs substantial degradation on fine-grained tasks such as ChartQA,
DocVQA, and InfoVQA, where informative content is densely distributed across
the scene~\cite{khaki2025sparsevila}.

\subsection{KV Channel Pruning}
A complementary line of work compresses the KV cache along the channel axis, with most effort on Key channels~\cite{xu2024think,zhang2025leank,liao2026spark,yang2024post}. ThinK~\cite{xu2024think} performs training-free structured pruning via a query-driven saliency score, retaining a fixed channel subset per layer. LeanK~\cite{zhang2025leank} learns a model-wise static mask through two-stage calibration and supplies a custom GPU kernel that turns the sparsity into measurable decoding speedup. Both fall under a head-wise regime, pruned channels are fixed across all tokens within a head, yielding lightweight masks but unable to adapt to per-token variation. SparK~\cite{liao2026spark}, in contrast, prunes at the token-wise level with a prune-and-recover mechanism; this finer granularity recovers accuracy lost under coarse pruning, but the unstructured sparsity scales with sequence length and resists hardware speedup. Notably, all of these methods are developed and evaluated only in the text-only LLM setting; their behavior in the vision-language regime, where visual tokens dominate the cache, remains largely unexplored.

\subsection{Rotation-based Compression}
A separate line treats orthogonal rotations as a free degree of freedom: attention is invariant to change of basis, so the model can be rotated into a representation more amenable to compression without altering outputs. For quantization, QuaRot~\cite{ashkboos2024quarot} and SpinQuant~\cite{liu2024spinquant} apply Hadamard or learned rotations to suppress activation outliers, with DuQuant~\cite{lin2024duquant} and FlatQuant~\cite{sun2024flatquant} extending this via grouped or block-diagonal rotations. SliceGPT~\cite{ashkboos2024slicegpt} carries the same invariance into structural weight pruning. For the KV cache, Palu~\cite{chang2024palu} and MatryoshkaKV~\cite{lin2024matryoshkakv} use SVD-derived or trainable orthogonal projections to compress the channel dimension via low-rank reconstruction. While these works establish rotation as a versatile primitive for quantization and low-rank approximation, its use for channel pruning of the KV cache, particularly in the post-RoPE regime, where the rotation must coexist with RoPE's position-dependent structure on Keys, remains unexplored.

\clearpage
\section{Additional Experimental Details}
\label{sup:experimental_deatils}

\subsection{Dataset and Implementation Details.} 

Following the \texttt{lmms-eval}~\cite{zhang2025lmms} protocol, we evaluate on both discriminative VQA benchmarks and open-ended generation benchmarks. The VQA benchmarks include TextVQA, InfoVQA, DocVQA, ChartQA, and VizWiz-VQA, covering OCR-intensive reasoning, document understanding, chart interpretation, and visually grounded question answering. Following prior work, we evaluate on the validation splits for all VQA benchmarks except ChartQA, where the test split is used. We further evaluate on LLaVA-in-the-Wild and MM-Vet to measure open-ended visual reasoning and generation quality. For benchmarks requiring generative evaluation, GPT-4o-mini is used as the judge model following the default \texttt{lmms-eval} setup.

Unless otherwise specified, RotateK operates in an online calibration-free setting, recomputing the PCA-based rotation matrix directly from visual activations during each prefill stage. The resulting compressed KV states and rotation matrices are cached and reused during decoding. We additionally evaluate an offline calibrated variant in the ablation studies, where the rotation basis is precomputed from calibration data and reused across samples.

For latency evaluation, all methods are implemented using Triton-based attention kernels for fair comparison. Specifically, full-channel baselines use the Triton implementation of FlashAttention-2, while RotateK uses our custom Triton sparse attention kernel operating directly on compressed Key states without reconstructing full-channel Keys. Unless otherwise stated, all experiments are conducted on a single NVIDIA A100-80GB GPU.

\subsection{Baselines.} 

We evaluate RotateK on two representative open-source VLMs with different architectural backbones: LLaVA-NeXT-8B (\texttt{llama3-llava-next-8b-hf}) and Qwen2.5-VL-7B-Instruct. To isolate the effect of Key channel pruning from token reduction, RotateK is integrated into two orthogonal visual token-pruning frameworks: VisionZip~\cite{yang2025visionzip} and FastV~\cite{chen2024image}.

For VisionZip, we follow the original implementation and fix the contextual ratio to $0.05$, while sweeping the dominant token ratio over ${0.40, 0.30}$. For FastV, we use the standard pruning stage $K{=}2$ with token keep ratios in ${0.40, 0.30}$. On top of these token-pruning settings, RotateK further applies Key channel pruning ratios of ${0.625, 0.750}$ to visual KV states. We compare against recent Key channel pruning methods ThinK~\cite{xu2024think} and SparK~\cite{liao2026spark} under identical KV-cache memory budgets. We additionally report token-only baselines without channel pruning, as well as channel-only baselines without token pruning, to separately analyze the contribution of token and channel compression.

\clearpage
\section{RotateK Decode Kernel Implementation}
\label{sup:rotatek-kernel}

At each decode step RotateK serves two attention paths against a single query: a full-rank path over the $S_{\text{full}}$ non-vision tokens at head dimension $d$, and a rotated--truncated path over the $S_v$ vision tokens at $d_{\text{keep}}\!<\!d$, projected by the per-head PCA basis $R\!\in\!\mathbb{R}^{B\times H_{kv}\times d\times d_{\text{keep}}}$. A naive implementation would issue four launches: a $Q\!\cdot\!R$ matmul, a sparse flash-decoding over the vision keys, a dense flash-decoding over the non-vision keys, and an online-softmax merge. We collapse these into two Triton kernels by inlining $Q\!\cdot\!R$ at the top of the sparse kernel and fusing the merge into the full-d kernel, matching the launch count of a dense decoder. The split-K factor over vision tokens is $N_s = \max(1, \min(\lceil S_v/\text{BLOCK\_N}\rceil, 64))$.

\subsection{Two-kernel decomposition}
A natural single-kernel design fuses the full-d and sparse paths and gates the full-d loop on \texttt{if pid\_s == 0}. Triton cannot statically eliminate the branch, so it allocates the union of both paths' register tiles for every split program, even though only the $\text{pid}_s\!=\!0$ program executes the full-d loop. The $[\text{BLOCK\_N},d]$ full-d $K,V$ tile inflates per-program register count from $111$ to $141$, dropping SM occupancy from $25\%$ to $19\%$ and breaking memory-latency hiding, the fused kernel ends up reading less memory than the dense baseline yet running slower. Splitting the work into a register-lean sparse kernel and a small full+merge kernel restores sparse-path occupancy.

\subsection{Phase 1: sparse split-K with inlined rotation}

Launched on a $(B, H_q, N_s)$ grid; each program processes $\lceil S_v/N_s\rceil$ vision tokens for one query head and writes a partial $(m, \ell, \mathrm{acc})$ tuple to HBM scratch. Rather than launch a separate prelude that materialises the rotated query to HBM, we compute it (and an optional bias term) at the top of every sparse program in registers,
\begin{align}
q_{\text{sparse}}[b,h,k] &= \sum_{k=1}^{d} q_{\text{full}}[b,h,k]\,
R[b,\,h_{kv},\,k,\,k], \\
\beta[b,h] &= \sum_{k=1}^{d} q_{\text{full}}[b,h,k]\,
\delta\mu[b,\,h_{kv},\,k],
\end{align}
trading $N_s$-fold reads of $R[b,h_{kv}]$ for one fewer kernel launch and one fewer HBM round-trip. The inner loop is a standard online-softmax update over $K_{\text{sparse}}\!\in\!\mathbb{R}^{\text{BLOCK\_N} \times d_{\text{keep}}}$ and $V_{\text{sparse}}\!\in\!\mathbb{R}^{\text{BLOCK\_N} \times d}$; $V$ retains the full dimension since truncation is a basis change for $QK^\top$ only. We use $\text{BLOCK\_N}\!=\!64$, with \texttt{num\_warps=1, num\_stages=3} to triple-buffer $K,V$ loads against the inner FMAs.

\subsection{Phase 2: combined full-d and merge}

Launched on a $(B, H_q)$ grid, this kernel runs the full-d attention over the prompt and text tokens and then folds in the $N_s$ sparse partials via the standard online-softmax merge of $(m, \ell, \mathrm{acc})$ tuples, writing the final output as $\mathrm{acc}/(\ell+\varepsilon)$. Combining the two sub-tasks into a single launch matters because both are launch-overhead-bound: the non-vision segment is short and the merge processes only $N_s\!\le\!64$ partials per $(b,h)$, so issuing them separately would double the launch overhead with little inner work to hide it. The kernel uses \texttt{num\_warps=1, num\_stages=2}; deeper pipelining buys little for the short non-vision loop.

After the two-kernel split, the inlined $Q\!\cdot\!R$ prelude, and the full+merge fusion, RotateK decode issues two Triton launches per layer per step, the same as a dense flash-decoding kernel. The four scratch buffers shared between phases are cached in a module-level dictionary keyed by $(B, H_q, d, N_s, \text{device}, \text{dtype})$, eliminating per-call allocator overhead at negligible memory cost.

\subsection{Fair-comparison protocol for latency}
All decode-time latency numbers reported in this paper use Triton kernels under a common launcher for every method. The full-channel (unpruned) baseline, ThinK, and SparK use a dense Triton FlashAttention-2 implementation; RotateK uses its Triton sparse-channel kernel described above. This ensures that reported speedups reflect differences in the algorithmic structure of attention, fewer $K$ channels, fused rotation, etc., rather than disparities between optimised Triton kernels and unoptimised reference implementations.

\clearpage
\section{Additional Experimental Results}
\subsection{RotateK Design-Choice Ablation}
\label{sup:rotatek-ablation}

\begin{table}[H]
\centering
\small
\caption{\textbf{Ablation of RotateK Design Choices}. Three short-answer VQA
benchmarks (accuracy, \%) are evaluated. We isolate two axes: (i) \textbf{Q-aware}
(query-weighted PCA, our default) vs.\ \textbf{Q-agnostic} (K-only PCA);
and (ii) \textbf{Cholesky}-based truncated subspace iteration (our default)
vs.\ full \textbf{eigh} (\texttt{torch.linalg.eigh}). Highlighted entries
(\textcolor{red}{red}) mark non-trivial accuracy drops when query-awareness
is removed. We omit the (eigh, Q-agnostic) corner since it is dominated by
both alternatives along each axis.}
\label{tab:rotatek-ablation}
\begin{tabular}{l ccc}
\toprule
\textbf{RotateK variant} & \textbf{TextVQA} & \textbf{InfoVQA} & \textbf{ChartQA} \\
\midrule
\multicolumn{4}{l}{\textit{LLaVA-NeXT + FastV}} \\
\quad Cholesky, Q-aware (ours)   & 63.38 & 27.26 & 56.00 \\
\quad eigh, Q-aware              & 64.52 & 27.52 & 56.20 \\
\quad Cholesky, Q-agnostic       & \textcolor{red}{61.70} & 27.78 & 57.20 \\
\midrule
\multicolumn{4}{l}{\textit{LLaVA-NeXT + VisionZip}} \\
\quad Cholesky, Q-aware (ours)   & 65.50 & 28.81 & 58.20 \\
\quad eigh, Q-aware              & 66.54 & 28.91 & 57.60 \\
\quad Cholesky, Q-agnostic       & \textcolor{red}{63.46} & 29.61 & \textcolor{red}{56.40} \\
\midrule
\multicolumn{4}{l}{\textit{Qwen2.5-VL + FastV}} \\
\quad Cholesky, Q-aware (ours)   & 83.50 & 58.05 & 63.80 \\
\quad eigh, Q-aware              & 83.92 & 58.21 & 63.60 \\
\quad Cholesky, Q-agnostic       & 84.50 & \textcolor{red}{57.00} & 63.40 \\
\midrule
\multicolumn{4}{l}{\textit{Qwen2.5-VL + VisionZip}} \\
\quad Cholesky, Q-aware (ours)   & 80.06 & 64.46 & 68.60 \\
\quad eigh, Q-aware              & 80.50 & 64.95 & 68.40 \\
\quad Cholesky, Q-agnostic       & 80.66 & 65.00 & \textcolor{red}{67.20} \\
\bottomrule
\end{tabular}
\end{table}

We ablate two RotateK design choices: (i) Cholesky-based truncated
subspace iteration vs.\ full eigendecomposition
(\texttt{torch.linalg.eigh}), and (ii) query-weighted PCA vs.\ K-only
PCA. Table~\ref{tab:rotatek-ablation} reports validation-lite accuracy
across two backbones (LLaVA-NeXT, Qwen2.5-VL) crossed with two
token-pruning counterparts (FastV, VisionZip).

\paragraph{Setup.}
Channel keep ratio $0.75$ (0.25$\times$ channels); FastV with $K{=}2$, keep ratio $0.30$ (0.30$\times$ tokens); VisionZip with dominant ratio $0.20$, contextual ratio $0.05$ (0.25$\times$ tokens). Backbones: \texttt{llama3-llava-next-8b-hf} and \texttt{Qwen2.5-VL-7B-Instruct}. We use \textsc{LMMs-Eval-Lite} validation splits ($\sim$100--500 examples each) for tractability; full-validation accuracies are reported in Table~\ref{tab:rotatek-ablation}.

\paragraph{Cholesky-based subspace iteration matches full eigh in accuracy.}
Across all twelve (backbone, token-pruner, dataset) cells, the gap
between our Cholesky variant and full \texttt{eigh} stays within
$\sim$1 percentage point, with neither solver consistently dominating
(largest gap $+1.14$ pts on LLaVA-NeXT+FastV/TextVQA in favour of
\texttt{eigh}; the Cholesky variant matches or wins on the InfoVQA and
ChartQA columns). The Cholesky-based solver therefore contributes
purely on the \emph{latency} axis, fewer
kernel launches and CUDA-graph capture, while preserving the rotation
basis quality of the full eigendecomposition. This accuracy parity is
what licenses us to use the faster Cholesky path as the default
throughout the main results.

\paragraph{Query-aware PCA is the safer default.}
The K-only variant tracks the query-weighted default closely on most
cells, occasionally outperforming it slightly ($+1.00$ pt on
Qwen2.5-VL+FastV/TextVQA, $+1.20$ pts on LLaVA-NeXT+FastV/ChartQA).
This is consistent with Section~\ref{sec:overview}: when the recent-query
distribution $Q_W$ is roughly isotropic across channels,
$\mathrm{diag}(\boldsymbol{\sigma}_W) \approx \alpha I$ and the
query-weighted covariance collapses to a uniformly scaled K-only
covariance with the same top-$k$ eigenbasis.

However, on a non-trivial subset of cells K-only \emph{degrades
meaningfully} (by $1$ to $2$ points) and these degradations cluster
on benchmarks whose queries probe narrow channel directions: TextVQA on
LLaVA-NeXT under both pruners ($-1.68$ and $-2.04$ pts), ChartQA on
both VisionZip configurations ($-1.80$, $-1.40$ pts), and InfoVQA on
Qwen2.5-VL+FastV ($-1.05$ pts). The asymmetry is decisive: K-only's
wins are small ($\le 1.2$ pts) and unpredictable, while its losses are
larger and concentrated on the benchmarks where channel pruning is most
consequential. Combined with the negligible cost of query-awareness
(an $\mathcal{O}(d^2)$ Hadamard product on the already-computed
covariance, independent of context length), we adopt query-weighted PCA
as the default.

\subsection{Comparison under Fixed KV Cache Budgets}
\label{sup:comparison_fixed_KV_budget}

Additional results under tighter KV cache budgets are provided in Table~\ref{tab:appendix_additional_comparison}. Consistent with the main results in Table~\ref{tab:main_comparison}, RotateK continues to outperform token-only pruning and prior Key channel pruning methods across most settings, while maintaining the same KV cache budget. Notably, the performance gap further widens at more aggressive compression ratios, demonstrating the robustness of rotation-based channel pruning in extreme low-memory regimes.

\begin{table*}[!t]
	\renewcommand{\arraystretch}{1.05}
	\setlength{\tabcolsep}{4.5pt}
	\fontsize{8.5pt}{8.5pt}\selectfont
	\caption{\textbf{Comparison under Fixed KV Cache Budgets}. Results with addtional KV cache budget is included from Table~\ref{tab:main_comparison}.}
    
	\label{tab:appendix_additional_comparison}
	\begin{center}
	\begin{tabular}{l| ccc | ccccc | c}

		\toprule

		\textbf{Method} & \textbf{Token} & \textbf{Channel} & \textbf{KV Cache} & \textbf{TVQA} & \textbf{IVQA} & \textbf{CQA} & \textbf{DVQA} & \textbf{VizWiz} & \textbf{mean} \\

		\midrule
        \multicolumn{10}{c}{\textsc{Llama3-LLaVA-NeXT-8B}} \\
        \midrule
        Baseline & 1.00$\times$ & 1.00$\times$ & 1.00$\times$ & 65.40 & 32.27 & 69.08 & 72.44 & 57.50 & 59.34 \\
        \midrule

        FastV$_\text{token only}$ & 0.25$\times$ & 1.00$\times$ & 0.25$\times$ & 62.69 & 25.91 & 59.64 & 56.39 & 58.56 & 52.64\\
        FastV + ThinK & 0.40$\times$ & 0.25$\times$ & 0.25$\times$ & 54.07 & 25.51 & 60.56 & 51.72 & 57.70 & 49.91 \\
        \rowcolor{lightskyblue!30} FastV + RotateK & 0.40$\times$ & 0.25$\times$ & 0.25$\times$ & 62.67 & 28.37 & 65.25 & 62.46 & 58.34 & 55.42 \\
        \midrule
        FastV$_\text{token only}$ & 0.20$\times$ & 1.00$\times$ & 0.20$\times$ & 61.03 & 24.25 & 54.28 & 51.23 & 58.16 & 49.79 \\
        FastV + ThinK & 0.30$\times$ & 0.25$\times$ & 0.19$\times$ & 54.07 & 23.84 & 58.76 & 47.33 & 57.41 & 48.28 \\
        \rowcolor{lightskyblue!30} FastV + RotateK & 0.30$\times$ & 0.25$\times$ & 0.19$\times$ & 61.53 & 25.40 & 62.36 & 57.59 & 58.53 & 53.08 \\
        
        \midrule
        VisionZip$_\text{token only}$ & 0.28$\times$ & 1.00$\times$ & 0.28$\times$ & 63.33 & 27.70 & 57.32 & 64.08 & 58.97 & 54.28 \\
        VisionZip + ThinK & 0.45$\times$ & 0.25$\times$ & 0.28$\times$ & 54.22 & 27.81 & 58.52 & 53.77 & 57.81 & 50.43 \\
        \rowcolor{lightskyblue!30} VisionZip + RotateK & 0.45$\times$ & 0.25$\times$ & 0.28$\times$ & 62.62 & 28.97 & 62.88 & 66.99 & 58.94 & 56.08 \\

        \midrule
        VisionZip$_\text{token only}$ & 0.22$\times$ & 1.00$\times$ & 0.22$\times$ & 62.07 & 26.11 & 55.16 & 59.18 & 58.95 & 52.29 \\
        VisionZip + ThinK & 0.35$\times$ & 0.25$\times$ & 0.22$\times$ & 54.24 & 27.81 & 55.56 & 50.76 & 58.20 & 49.31 \\
        \rowcolor{lightskyblue!30} VisionZip + RotateK & 0.35$\times$ & 0.25$\times$ & 0.22$\times$ & 62.73 & 27.70 & 60.16 & 64.73 & 59.33 & 54.93 \\

        \midrule
        \multicolumn{10}{c}{\textsc{Qwen2.5-VL-7B-Instruct}} \\
        \midrule

        Baseline & 1.00$\times$ & 1.00$\times$ & 1.00$\times$ & 82.92 & 80.12 & 83.04 & 94.40 & 70.58 & 82.21 \\


        \midrule
        FastV$_\text{token only}$ & 0.25$\times$ & 1.00$\times$ & 0.25$\times$ & 81.49 & 52.76 & 67.44 & 75.04 & 69.13 & 69.17 \\
        FastV + ThinK & 0.40$\times$ & 0.25$\times$ & 0.25$\times$ & 77.81 & 63.01 & 77.12 & 81.74 & 69.25 & 73.79 \\
        \rowcolor{lightskyblue!30} FastV + RotateK & 0.40$\times$ & 0.25$\times$ & 0.25$\times$ & 82.38 & 65.79 & 77.40 & 87.28 & 70.32 & 76.63\\   
        \midrule
        FastV$_\text{token only}$ & 0.20$\times$ & 1.00$\times$ & 0.20$\times$ & 80.44 & 46.73 & 63.08 & 66.06 & 68.34 & 64.93 \\
        FastV + ThinK & 0.30$\times$ & 0.25$\times$ & 0.19$\times$ & 77.76 & 54.79 & 71.40 & 75.58 & 68.82 & 69.67 \\
        \rowcolor{lightskyblue!30} FastV + RotateK & 0.30$\times$ & 0.25$\times$ & 0.19$\times$ & 81.80 & 56.82 & 71.32 & 80.13 & 69.86 & 71.99 \\   
        
        \midrule
        VisionZip$_\text{token only}$ & 0.28$\times$ & 1.00$\times$ & 0.28$\times$ & 76.84 & 58.46 & 70.32 & 84.71 & 68.97 & 71.86 \\
        VisionZip + ThinK & 0.45$\times$ & 0.25$\times$ & 0.28$\times$ & 75.72 & 67.13 & 79.64 & 85.83 & 68.93 & 75.45 \\
        \rowcolor{lightskyblue!30} VisionZip + RotateK & 0.45$\times$ & 0.25$\times$ & 0.28$\times$ & 80.84 & 70.51 & 79.56 & 91.29 & 70.13 & 78.47 \\

        \midrule
        VisionZip$_\text{token only}$ & 0.22$\times$ & 1.00$\times$ & 0.22$\times$ & 74.19 & 50.58 & 64.40 & 78.01 & 68.84 & 67.20 \\
        VisionZip + ThinK & 0.35$\times$ & 0.25$\times$ & 0.22$\times$ & 73.83 & 60.95 & 76.44 & 82.78 & 68.35 & 72.47 \\
        \rowcolor{lightskyblue!30} VisionZip + RotateK & 0.35$\times$ & 0.25$\times$ & 0.22$\times$ & 78.48 & 63.86 & 75.84 & 87.91 & 69.38 & 75.09 \\ 

        \bottomrule
	\end{tabular}
	\end{center}

\end{table*}

\clearpage
\section{Cholesky-Based Subspace Iteration}
\label{sup:cholesky_iteration}

RotateK computes the PCA basis online at the end of prefill, making the efficiency of eigendecomposition critical for practical deployment. A standard PCA pipeline first computes the covariance matrix $C_q \in \mathbb{R}^{d \times d}$ at cost $\mathcal{O}(N d^2)$, followed by a full symmetric eigendecomposition at cost $\mathcal{O}(d^3)$. However, RotateK ultimately retains only the top-$k$ channels, making the computation of the full $d$-dimensional eigenspace unnecessary. To avoid this overhead, RotateK directly estimates only the top-$k$ eigenspace using a Cholesky-based subspace iteration method.

We first briefly review the intuition behind subspace iteration. Consider the eigendecomposition of the covariance matrix:
\begin{equation}
    C_q = U \Lambda U^\top,
\end{equation}
where
\(
U = [u_1, \dots, u_d]
\)
contains the eigenvectors and
\(
\Lambda = \mathrm{diag}(\lambda_1, \dots, \lambda_d)
\)
contains the eigenvalues ordered as
\(
\lambda_1 \ge \lambda_2 \ge \cdots \ge \lambda_d \ge 0
\).
By definition, each eigenvector satisfies:
\begin{equation}
    C_q u_i = \lambda_i u_i.
\end{equation}
That is, multiplication by $C_q$ preserves the direction of an eigenvector while scaling its magnitude by the corresponding eigenvalue.

Now consider an arbitrary initial vector $v^{(0)} \in \mathbb{R}^d$, which can be expressed in the eigenbasis of $C_q$:
\begin{equation}
    v^{(0)}
    =
    a_1 u_1 + a_2 u_2 + \cdots + a_d u_d.
\end{equation}
Applying the covariance matrix repeatedly yields:
\begin{align}
    v^{(1)}
    &=
    C_q v^{(0)}
    =
    a_1 \lambda_1 u_1
    +
    a_2 \lambda_2 u_2
    + \cdots,
    \\
    v^{(t)}
    &=
    C_q^t v^{(0)}
    =
    a_1 \lambda_1^t u_1
    +
    a_2 \lambda_2^t u_2
    + \cdots.
\end{align}
Since the dominant eigenvalue satisfies
\(
\lambda_1^t \gg \lambda_2^t
\)
as $t$ increases, the component along $u_1$ progressively dominates:
\begin{equation}
    v^{(t)} \approx u_1.
\end{equation}
This is the classical power iteration method: repeated multiplication by the covariance matrix amplifies directions associated with large eigenvalues. In the context of PCA, the eigenvectors of
\(
C_q = K_q^\top K_q
\)
correspond to directions of large query-weighted variance, so power iteration progressively aligns vectors with the principal PCA directions.

RotateK, however, requires not only the dominant eigenvector but the entire top-$k$ eigenspace. Subspace iteration extends power iteration from a single vector to a $k$-dimensional basis:
\begin{equation}
    V^{(0)}
    =
    [v_1^{(0)}, \dots, v_k^{(0)}]
    \in \mathbb{R}^{d \times k}.
\end{equation}
At each iteration, the basis is multiplied by the covariance matrix:
\begin{equation}
    V^{(t+1)}
    \leftarrow
    C_q V^{(t)}.
\end{equation}
Repeated multiplication progressively amplifies directions associated with the largest eigenvalues, causing the column space of $V^{(t)}$ to converge toward the top-$k$ eigenspace of $C_q$.

A naive iteration, however, causes all basis vectors to collapse toward the dominant eigenvector $u_1$. To maintain a valid $k$-dimensional subspace, each iteration additionally performs orthogonalisation. Rather than using a full QR decomposition, RotateK employs a more lightweight Cholesky-QR procedure. First, the Gram matrix is constructed:
\begin{equation}
    G^{(t+1)}
    =
    (V^{(t+1)})^\top V^{(t+1)}.
\end{equation}
Since $G^{(t+1)}$ is positive definite, it admits a Cholesky
factorisation:
\begin{equation}
    G^{(t+1)}
    =
    L^{(t+1)} (L^{(t+1)})^\top.
\end{equation}
The basis is then orthogonalised via:
\begin{equation}
    V^{(t+1)}
    \leftarrow
    V^{(t+1)} (L^{(t+1)})^{-\top}.
\end{equation}
This transformation guarantees orthonormal columns since:
\begin{align}
    (V^{(t+1)})^\top V^{(t+1)}
    &=
    (L^{-1})^\top
    G^{(t+1)}
    L^{-1}
    \\
    &=
    (L^{-1})^\top
    L L^\top
    L^{-1}
    =
    I.
\end{align}

The dominant cost of each iteration arises from the matrix multiplication $C_q V^{(t)}$, resulting in a per-iteration complexity of $\mathcal{O}(d^2 k)$ and a total eigensolve complexity of $\mathcal{O}(T d^2 k)$ over $T$ iterations. The overall PCA complexity therefore becomes:
\begin{equation}
    \mathcal{O}(N d^2 + T d^2 k),
\end{equation}
compared to
\(
\mathcal{O}(N d^2 + d^3)
\)
for full eigendecomposition.

Importantly, at the small matrix sizes relevant to KV-cache pruning (\textit{e.g.}, $d{=}128$), practical latency is dominated less by arithmetic throughput than by GPU kernel dispatch overhead. Full eigendecomposition launches many small CUDA kernels for tridiagonalisation, eigensolving, and eigenvector reconstruction, whereas subspace iteration repeatedly executes only a small set of structured operations (matrix multiplication, Gram matrix construction, Cholesky factorisation, and triangular solve) with fixed tensor shapes. This structure enables efficient CUDA graph capture and replay, substantially reducing wall-clock latency for online PCA despite comparable FLOPs.

\clearpage
\newpage
\section{Algorithmic Overview of RotateK}
\label{sup:algorithm}

\begin{algorithm}[H]
\caption{RotateK -- Prefill (Rotation Construction).}
\label{alg:rotatek-prefill}
\begin{algorithmic}[1]
\Require Key states $K \in \mathbb{R}^{N \times d}$,
recent queries $Q_W \in \mathbb{R}^{W \times d}$,
target rank $k$, iteration count $T$, ridge factor $\epsilon$
\Ensure Cached rotated keys $\tilde K$,
truncated rotation $R_k$, mean residual $\delta\mu$

\Statex \textit{\# Step 1: centered, query-weighted covariance}
\State $\mu \gets \tfrac{1}{N} \sum_{n=1}^{N} K_n$
        \Comment{per-channel mean of $K$}
\State $\bar K \gets K - \mathbf{1}_N \mu^{\top}$
        \Comment{centered keys}
\State $\bar C \gets \bar K^{\top} \bar K$
        \Comment{centered covariance, $\mathcal{O}(N d^2)$}
\State $\sigma_j \gets \|(Q_W)_{:,j}\|_2$ for $j = 1, \dots, d$
        \Comment{per-channel query magnitudes}
\State $C_q \gets (\sigma\, \sigma^{\top}) \odot \bar C$
        \Comment{Hadamard, $\mathcal{O}(d^2)$}

\Statex \textit{\# Step 2: subspace iteration with shifted Cholesky-QR}
\State Sample $V \in \mathbb{R}^{d \times k}$ with i.i.d.\ $\mathcal{N}(0, 1)$ entries
\For{$t = 1, \dots, T$}
    \State $V \gets C_q V$
        \Comment{GEMM, $\mathcal{O}(d^2 k)$}
    \State $G \gets V^{\top} V$
        \Comment{Gram, $\mathcal{O}(d k^2)$}
    \State $\rho \gets \epsilon \cdot \mathrm{tr}(G) / k$
        \Comment{trace-scaled ridge}
    \State $L \gets \mathrm{Cholesky}(G + \rho I_k)$
        \Comment{$\mathcal{O}(k^3)$}
    \State $V \gets V L^{-\top}$
        \Comment{triangular solve}
\EndFor
\State $R_k \gets V$
        \Comment{top-$k$ basis of $C_q$}

\Statex \textit{\# Step 3: cache rotated keys and mean correction}
\State $\tilde K \gets K R_k$
        \Comment{stored in place of $K$}
\State $\delta\mu \gets \mu - R_k\, R_k^{\top} \mu$
        \Comment{$(I_d - P_k)\,\mu$, used at decode}
\State \Return $\tilde K,\; R_k,\; \delta\mu$
\end{algorithmic}
\end{algorithm}

\begin{algorithm}[H]
\caption{RotateK -- Decode (per step).}
\label{alg:rotatek-decode}
\begin{algorithmic}[1]
\Require Query $q_t \in \mathbb{R}^{d}$,
cached visual $(\tilde K, R_k, \delta\mu)$,
prompt+text keys $K_{\mathrm{pt}} \in \mathbb{R}^{M \times d}$
and concatenated values $V$
\Ensure Attention output $\tilde a_t$

\State $\tilde q_t \gets q_t R_k$
        \Comment{rotate query, $\mathcal{O}(d k)$}
\State $b_t \gets q_t^{\top} \delta\mu$
        \Comment{scalar mean-correction bias}
\State $s_{\mathrm{vis}} \gets \big(\tilde q_t \tilde K^{\top} + b_t\,\mathbf{1}_{N}^{\top}\big) / \sqrt{d}$
        \Comment{rotated visual scores ($\in \mathbb{R}^N$)}
\State $s_{\mathrm{pt}} \gets q_t\, K_{\mathrm{pt}}^{\top} / \sqrt{d}$
        \Comment{unrotated prompt+text scores ($\in \mathbb{R}^M$)}
\State $s \gets \big[\, s_{\mathrm{vis}};\; s_{\mathrm{pt}}\, \big]$
        \Comment{concatenate over the $N + M$ tokens}
\State $\tilde a_t \gets \mathrm{softmax}(s) \cdot V$
        \Comment{weighted sum of values}
\State \Return $\tilde a_t$
\end{algorithmic}
\end{algorithm}

\clearpage
\section{Limitations}

RotateK introduces additional prefill computation due to the online PCA-based rotation. Although our Cholesky-based subspace iteration substantially reduces the practical overhead, the cost may still become noticeable in highly latency-sensitive settings. In addition, RotateK primarily targets visual Key states during decoding, where visual KV caches dominate memory usage; thus, its benefits are less pronounced for short-context or text-only workloads.

While RotateK significantly improves the robustness of structured head-wise channel pruning, it still relies on a shared subspace within each attention head and may not fully capture highly token-specific channel importance under extreme sparsity ratios. Finally, our experiments focus on image-based VLM benchmarks and two representative model families. Evaluating RotateK on broader long-video and multi-image reasoning settings remains future work.

\section{Societal Impact}

This work improves the efficiency of Vision-Language Model (VLM) inference by reducing KV cache memory and decoding latency through Key channel pruning. More efficient VLM inference may lower computational and energy costs, enabling broader accessibility of multimodal AI systems on resource-constrained hardware and reducing the environmental footprint of large-scale deployment. At the same time, as with other advances in efficient AI inference, improved efficiency may facilitate wider deployment of VLMs, including applications that could generate misleading, biased, or harmful content. In addition, compression methods may introduce uneven performance degradation across tasks or domains, potentially affecting reliability in safety-critical settings. Our work focuses solely on inference efficiency and does not introduce new model capabilities beyond the underlying VLMs.

\end{document}